\pdfoutput=1

\documentclass[11pt]{article}

\usepackage[final]{acl}

\usepackage{times}
\usepackage{latexsym}

\usepackage[T1]{fontenc}

\usepackage[utf8]{inputenc}

\usepackage{microtype}

\usepackage{inconsolata}

\usepackage{mathrsfs}
\usepackage{amsmath,amssymb,mathtools}
\usepackage{booktabs}
\usepackage{multirow}
\usepackage{dcolumn}
\usepackage{comment}

\title{The Curse of Popularity:\\Popular Entities have Catastrophic Side Effects when \\ Deleting Knowledge from Language Models}

\author{
    Ryosuke Takahashi${}^{1}$\quad
    Go Kamoda${}^{1}$\quad 
    \\
    {\bf Benjamin Heinzerling${}^{2,1}$\quad
    Keisuke Sakaguchi${}^{1,2}$\quad
    Kentaro Inui${}^{3,1,2}$}
    \\
    ${}^{1}$Tohoku University\quad
    ${}^{2}$RIKEN\quad
    ${}^{3}$MBZUAI\\
    \texttt{\{ryosuke.takahashi, go.kamoda\}@dc.tohoku.ac.jp}\\
    \texttt{benjamin.heinzerling@riken.jp}\quad
    \texttt{keisuke.sakaguchi@tohoku.ac.jp}\\ \texttt{kentaro.inui@mbzuai.ac.ae}
}

\begin{document}
\maketitle
\begin{abstract}
Language models (LMs) encode world knowledge in their internal parameters through training.
However, LMs may learn personal and confidential information from the training data, leading to privacy concerns such as data leakage.
Therefore, research on knowledge deletion from LMs is essential.
This study focuses on the knowledge stored in LMs and analyzes the relationship between the side effects of knowledge deletion and the entities related to the knowledge. 
Our findings reveal that deleting knowledge related to popular entities can have catastrophic side effects.
Furthermore, this research is the first to analyze knowledge deletion in models trained on synthetic knowledge graphs, indicating a new direction for controlled experiments.
\end{abstract}

\section{Introduction}
\label{sec:introduction}
Language models (LMs) can store knowledge in their internal parameters through training~\citep{petroni-etal-2019-language}, and research focusing on analyzing the knowledge stored inside LMs has gained attention~\citep{jiang-etal-2020-know, heinzerling-inui-2021-language}.
Although this capability is essential in building human-aiding assistants, challenges related to reliability and safety are also reported. 
For example, LMs possess knowledge only up to the point when their training data was collected, making them not robust to the constantly changing real-world knowledge~\citep{Kasai2022-lw}.
Additionally, there is a risk that LMs could leak personal and confidential information contained in the training data, raising privacy concerns~\citep{huang-etal-2022-large}.
To address these challenges, several studies have been conducted on \emph{knowledge editing}~\citep{Feng2023-ew, Zhang2024-jk, dai-etal-2022-knowledge, ROME, meng2022memit, li2023pmet} and \emph{knowledge deletion}~\citep{jang-etal-2023-knowledge, Ishibashi2023-ng} in LMs.
These studies have reported some success in knowledge editing and deletion, yet failures, challenges, and difficulties have also been identified.

\begin{figure*}[t]
\centering
\includegraphics[width=\linewidth]{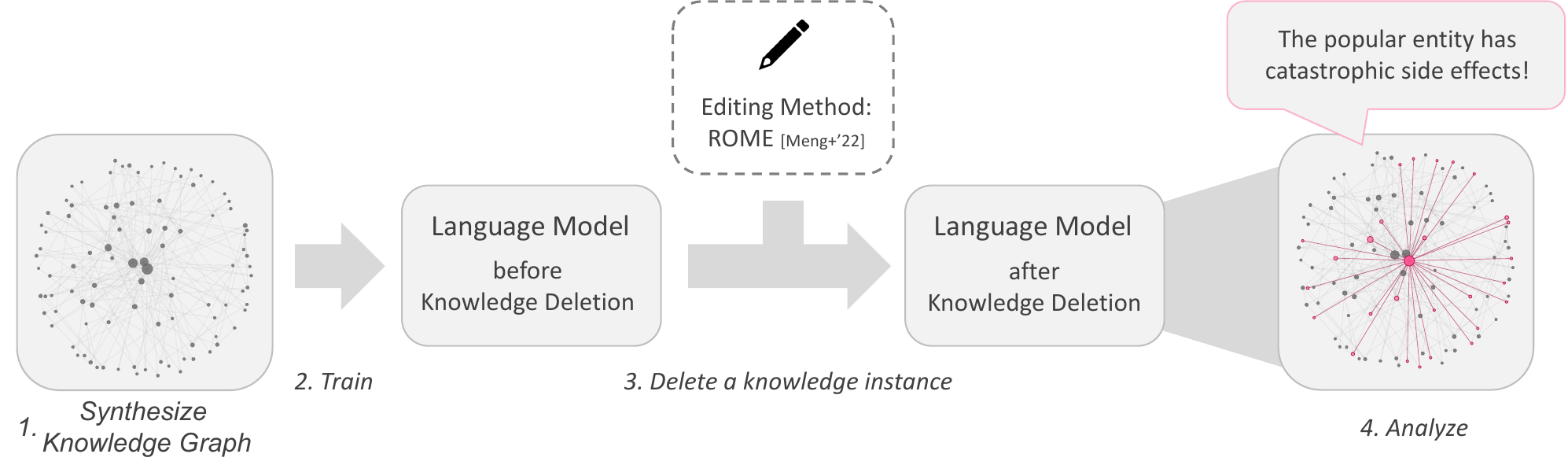}
\caption{
Overview of the analysis flow for the side effects of knowledge deletion using a synthetic knowledge graph:
\textit{1.} First, create a synthetic knowledge graph.
\textit{2.} Train the LM on the created knowledge graph.
\textit{3.} Apply the knowledge editing method, ROME, to delete a specific knowledge instance.
\textit{4.} Analyze the side effects of the deleted knowledge by comparing the model's accuracy on the trained knowledge before and after the deletion.
As a result, we reveal that deleting knowledge related to popular entities has catastrophic side effects.
}
\label{fig:fig1}
\end{figure*}

Our study aims to understand better when and why knowledge editing and deletion work as intended.
We hypothesize that analysis of what kind of knowledge is being deleted is essential to fulfill the objective.
This paper focuses on the frequency of entities related to knowledge in the training corpus.
To test this hypothesis, we design controlled experiments to analyze the impact of knowledge deletion on LMs (\autoref{fig:fig1}).
We formalize the notion of ``kind of knowledge'' regarding the structural properties of the knowledge graph trained by the model. 
We synthesize knowledge graphs with various properties, pre-train LMs on these graphs, perform knowledge deletion of specific facts, and then observe the side effects of deletion on entities related to the deleted knowledge.

Analyses reveal that deleting knowledge concerning frequently occurring entities results in significant and catastrophic side effects in LMs trained on knowledge graphs with similar properties to the real world.
Furthermore, this study is the first to analyze knowledge deletion using models with controlled knowledge structures, presenting a new direction in the analysis of knowledge deletion using synthetic knowledge graphs.

\section{Experimental Design and Approach}
\label{sec:design_approach}
LMs distribute and encode knowledge across many model parameters, forming complex interdependencies among various knowledge instances.
This interconnected structure may lead to unintended side effects when deleting specific knowledge.
Furthermore, not all knowledge is equally essential; ``more important'' knowledge might be connected to many others, and ``less important'' knowledge with fewer associations.
However, accurately identifying what knowledge has been stored in pre-trained LMs is challenging~\citep{jiang-etal-2020-know}, making it difficult to conduct detailed analyses considering the properties of the knowledge structure held internally by existing pre-trained LMs.
In light of this, we have designed the following experiment.

First, we create a synthetic knowledge graph (\autoref{sec:knowledge-graph}) and train the LM on the created knowledge graph (\autoref{sec:training}).
This enables us to control the knowledge stored in the LM through training precisely.
Then, we apply the knowledge editing technique to the LM to delete a specific knowledge instance (Sections \ref{sec:ROME} and \ref{sec:knowledge-deletion}) and analyze side effects to test our hypotheses (\autoref{sec:experiment}).
This experimental design allows us to conduct analyses focusing on the properties of the knowledge structure concerning knowledge deletion.

\section{Experimental Setup}
\label{sec:experimental-setup}

\subsection{Knowledge Graphs}
\label{sec:knowledge-graph}
This work deals with relational knowledge represented in triples, such as ($s$, $r$, $o$).
We refer to a representation where the subject $s$ and object $o$ correspond to vertices, and the relation $r$ corresponds to an edge, as a knowledge graph.
Here, $s$ and $o$ are elements of the entity set $\mathcal{E}$ with $|\mathcal{E}| = 200$, and $r$ is an element of the relation set $\mathcal{R}$ with $|\mathcal{R}| = 50$.

\begin{figure}[t]
\centering
\includegraphics[width=\linewidth]{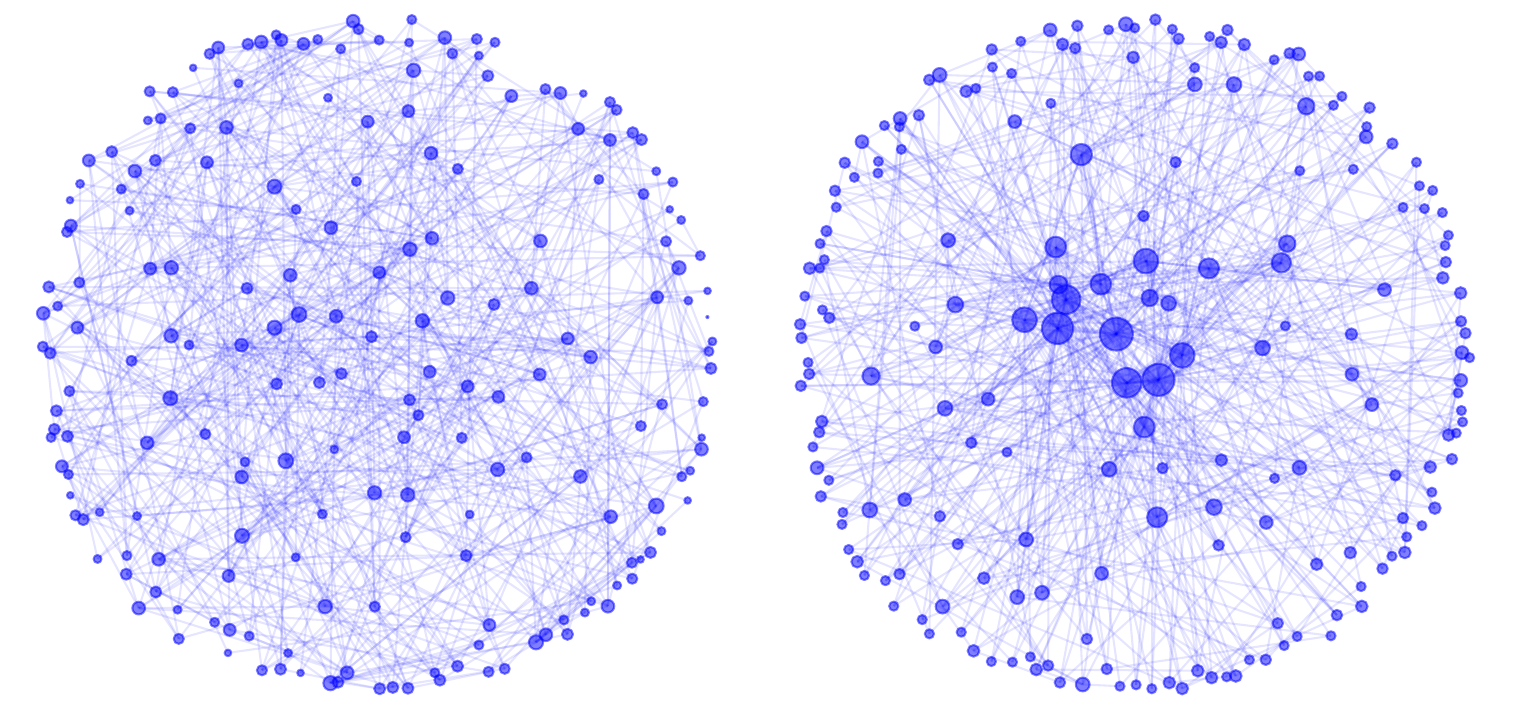}
\caption{
Two synthetic knowledge graphs we created.
(Left) Erdős-Rényi graph: Features a relatively uniform degree distribution of vertices, representing a simple structure.
(Right) Barabási-Albert graph: The degree distribution of vertices follows a power law, reflecting the properties of complex networks in the real world.
}
\label{fig:graph}
\end{figure}

Previous studies on knowledge editing and deletion have assumed the existence of knowledge graphs composed of relational knowledge between entities expressed in natural language.
This study introduces an innovative approach by creating a synthetic knowledge graph, allowing for precise control over the information LMs acquire during training. 
We created two synthetic knowledge graphs with different characteristics.
The first is an Erdős-Rényi (ER) graph~\citep{erdos59a}, structured to ensure the probability of forming edges between vertices is uniform.
The second is a Barabási-Albert (BA) graph~\citep{barabasi1999emergence}, characterized by a vertex degree distribution that follows a power law, thus resembling the structure of the real world more closely (\autoref{fig:graph}).
One advantage of synthetic knowledge graphs is that, for instance, it is possible to create entities with extremely high or low degrees, enabling the analysis of knowledge deletion for such entities in the BA graph. 

\subsection{Storing Knowledge Graphs in LMs}
\label{sec:training}
We train LMs using synthetic knowledge graphs.
With the knowledge graphs created in \autoref{sec:knowledge-graph}, we first assign five names to each entity $e^i$ $(0 \leq i < |\mathcal{E}| \, ; i \in \mathbb{N})$, denoted as $e^i_j$ $(0 \leq j < 5 \, ; j \in \mathbb{N})$.
Similarly, we assign five names to each relation $r^i$ $(0 \leq i < |\mathcal{R}| \, ; i \in \mathbb{N})$, denoted as $r^i_j$ $(0 \leq j < 5 \, ; j \in \mathbb{N})$.
The LMs include these names in their vocabulary.
Hereafter, names referring to the same entity (or relation) are termed ``paraphrases.''
We then create a corpus composed of sentences with three words (e.g., ``$e^0_0$ $r^1_1$ $e^1_{4}$'') and train models with GPT-2~\citep{radford2019language} architecture, initialized with 6, 12, and 24 layers.

\begin{figure}[t]
\centering
\includegraphics[width=\linewidth]{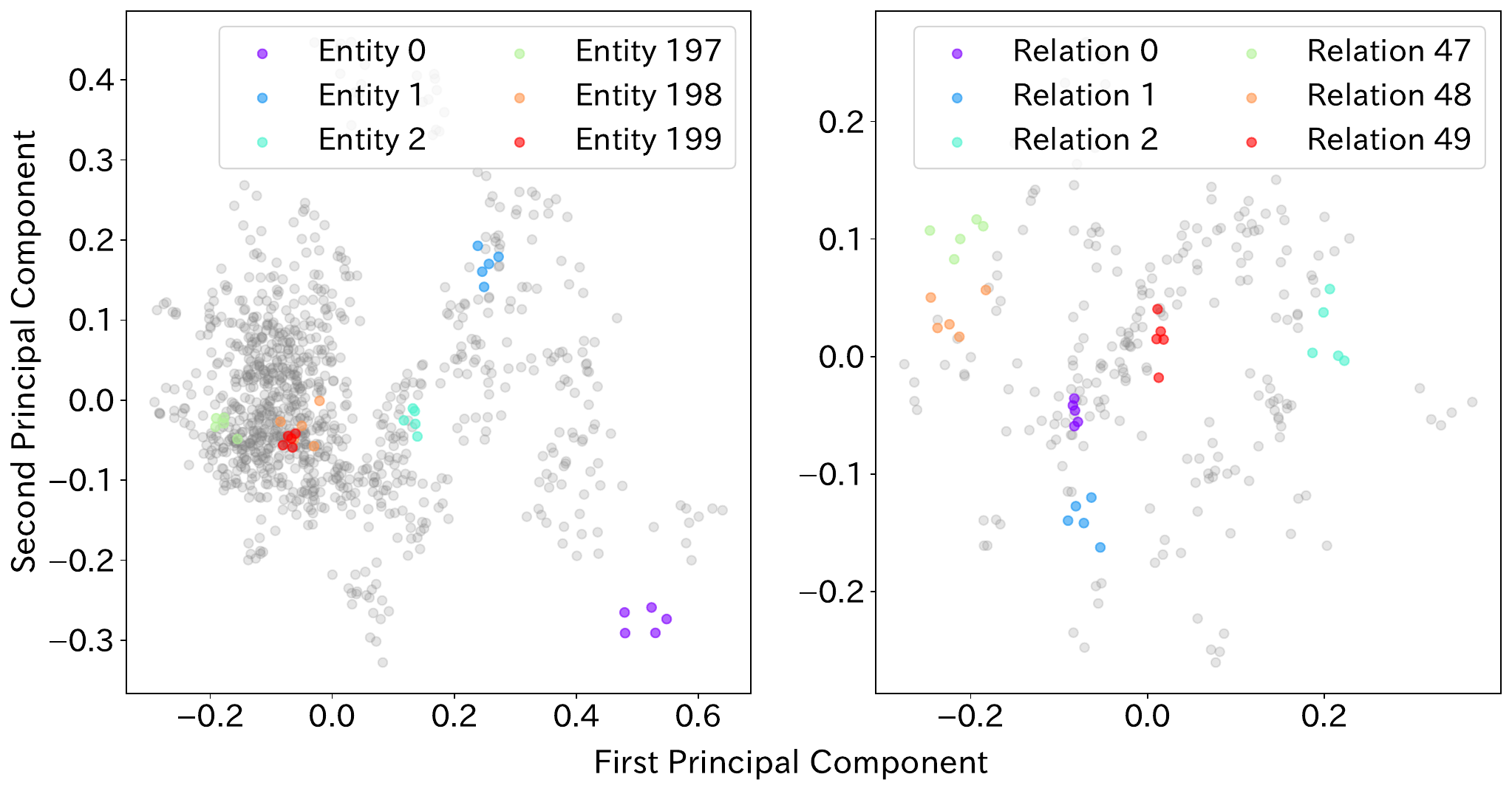}
\caption{
Results of the principal component analysis (PCA) on the embedding representations in the LM after training ER graphs.
The left side represents the PCA results for entity embeddings, while the right represents the PCA results for relation embeddings. 
Each entity and relation has five paraphrases and paraphrases about the same entity or relation are illustrated in the same color (here, paraphrases of six entities and relations are highlighted). 
The PCA results indicate that the embeddings of paraphrases cluster together, suggesting that the LM recognizes paraphrases.
}
\label{fig:embeddings}
\end{figure}

During inference, we input two words into the model, and the model predicts one word.
A prediction is correct if the outputted word represents any paraphrases indicating the gold entity.
During training, we use $20\%$ of the entire knowledge base, which includes paraphrased knowledge, intending to achieve generalization across all paraphrased knowledge (for further details, refer to \autoref{sec:paraphrase}).
Then, we use the entire knowledge as full data to verify whether the model has successfully learned the knowledge graph.
After training, the model achieves an accuracy rate of approximately $99\%$ not only on the training data but also on the full data, indicating that it has successfully memorized the provided knowledge (see \autoref{tab:acc}).
Additionally, principal component analysis on the embeddings of entities and relations in the trained model suggests that the model recognizes the representations of paraphrases (\autoref{fig:embeddings}).

\begin{table}[t]
\centering
\small
\begin{tabular}{cccc}
\toprule
\begin{tabular}[c]{@{}c@{}}Knowledge\\Graph\end{tabular} & Model & \begin{tabular}[c]{@{}c@{}}Accuracy on\\ Training Data\end{tabular} & \begin{tabular}[c]{@{}c@{}}Accuracy on\\ Full Data\end{tabular} \\ \midrule\midrule
\multirow{3}{*}{ER} & \phantom{0}6 layers & .9998 & .9941 \\
 & 12 layers & .9985 & .9983 \\
 & 24 layers & .9900 & .9905 \\ \midrule
\multirow{3}{*}{BA} & \phantom{0}6 layers & .9991 & .9987 \\
 & 12 layers & .9991 & .9989 \\
 & 24 layers & .9972 & .9964
\\ 
\bottomrule
\end{tabular}
\caption{
Accuracy of training data and full data for each model after training the RA graph and the BA graph.
}
\label{tab:acc}
\end{table}

\subsection{Knowledge Editing}
\label{sec:ROME}
While our experimental design is agnostic to knowledge editing and deletion methods, here we opt for ROME~\citep{ROME}, one of the primary knowledge editing methods for causal LMs.
ROME updates model weight through the following two steps to achieve the editing of a specific knowledge instance.

\paragraph{Step 1: Causal Tracing} 
First, we identify the model components that play a crucial role in the process of knowledge association by the model.
This is achieved by analyzing the contribution of each hidden state of the model when predicting information related to knowledge\footnote{Here, predicting $o$ for an input $x=(s, r)$ is referred to.} (for details, see \autoref{sec:causal_tracing}).
The results of Causal Tracing revealed that the initial Feed-Forward (FF) layers significantly contribute to knowledge association, supporting research indicating that FF layers serve as key-value memory storage for knowledge~\citep{geva-etal-2021-transformer}.

\paragraph{Step 2: Rank-One Model Editing} 
During editing, we add a rank-1 matrix to the weights $W_2$ of the second layer in the FF layer (\autoref{eq:ffn}) identified by Causal Tracing.
\begin{align}
\label{eq:ffn}
    \text{FFN}(x) = \sigma(xW_1 + b_1)W_2 + b_2
\end{align}
Specifically, we treat $W_2$ as an associative memory for existing key-value pairs $(K, V)$, and perform edits to insert a new key-value pair $(k_*, v_*)$\footnote{$K=[k_1|k_2|\ldots]$ and $V=[v_1|v_2|\ldots]$, where $k$ and $v$ represent vectors.}.
This reduces to solve a constrained least-squares problem, and the updated weight $\hat{W}_2$ is given as follows.
\begin{align}
    \hat{W}_2 = W_2 + \Lambda(C^{-1}k_*)^\top
\end{align}
Here, $C^{-1}=KK^\top$ represents the uncentered covariance of $K$, and $\Lambda$ is a vector proportional to the residual error of the new key-value pair $(k_*, v_*)$.

\subsection{Knowledge Deletion from LMs}
\label{sec:knowledge-deletion}

Generally, knowledge editing in LMs refers to reassociating an entity with a different entity from one already connected to.
This means updating a knowledge instance from $(s, r, o)$ to $(s, r, o^*)$.
In this work, we define knowledge deletion as the process of reassociating an entity with a different entity from one it is already connected to, by introducing an entity $e_\text{deleted}$ to represent deletion.
By treating $e_\text{deleted}$ entity as $o^*$, we achieve knowledge deletion using a pre-existing knowledge editing method.

\section{Experiment}
\label{sec:experiment}

\paragraph{Procedure}
We analyze the side effects of deleting a specific knowledge instance.
This means measuring the impact of deleting a knowledge instance on the overall set of knowledge instances.
The impact is determined as follows:

\begin{enumerate}
    \item Measure the accuracy of knowledge at the end of pre-training, denoted as $\text{acc}_\text{pre-del}$.

    \item Delete one piece of knowledge related to a subject $s$. 
    This means editing the knowledge from $(s, r, o)$ to $(s, r, e_\text{deleted})$ to delete the knowledge instance.    

    \item After deleting the knowledge instance related to the subject $s$, measure the accuracy, $\text{acc}_\text{post-del}(s)$, for all knowledge instances in the model.

    \item We calculate the impact $I(s)$ as the difference in accuracy before and after deleting a knowledge instance related to the subject $s$.
    \begin{align}
    I(s) = \text{acc}_\text{pre-del} - \text{acc}_\text{post-del}(s)
    \end{align}

\end{enumerate}

\begin{figure*}[t]
\centering
\includegraphics[width=\linewidth]{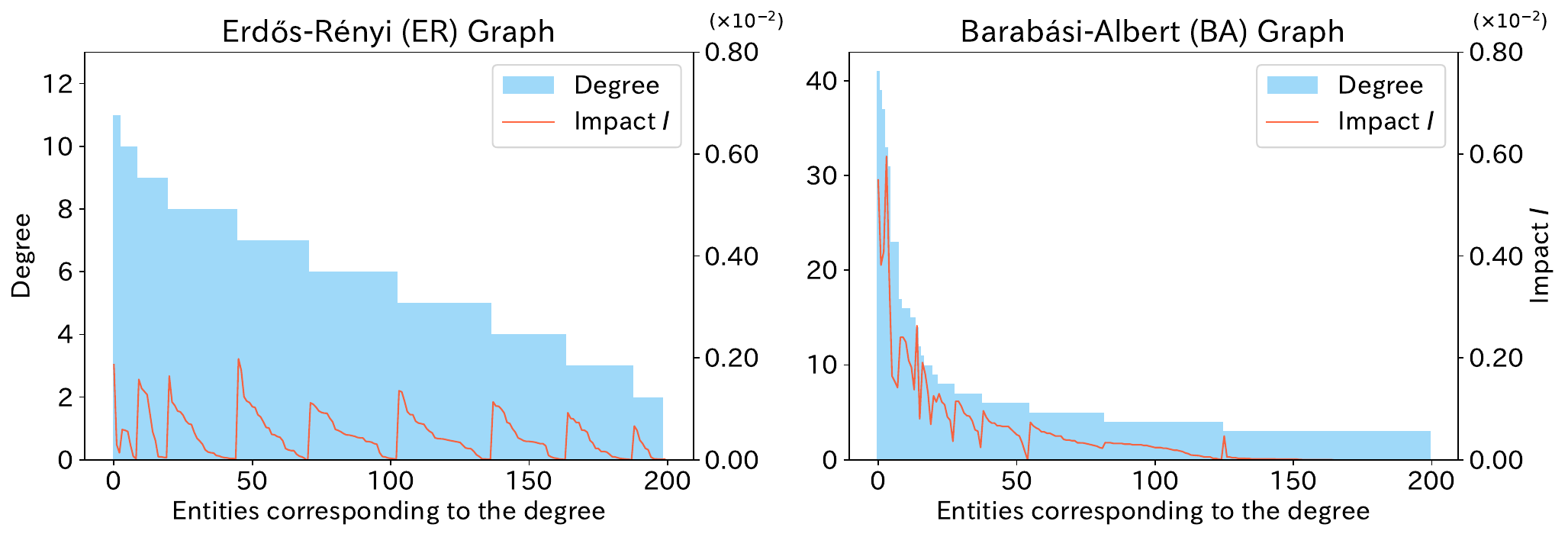}
\caption{
The relationship between the degree of entities, or subjects, and the impact of their deletion in a 6-layer GPT model trained on the knowledge graph.
The left vertical axis indicates the degree of entities, while the horizontal axis represents the corresponding entities.
The right vertical axis shows the impact (amount of side effects) on other knowledge when deleting a knowledge instance related to an entity.
When it comes to deleting knowledge related to a specific entity, we have observed that there is no relationship between the degree (i.e., number of connections) of the entity in the ER graph and the impact of its deletion. However, in the BA graph, there is a clear relationship between the degree of the entity and the impact of its deletion.
Since the impact of deleting knowledge in LMs can have significant side effects on related knowledge, it is recommended to avoid deleting knowledge related to frequent entities in LMs trained on knowledge structures that are closer to the real world, as doing so may have catastrophic consequences.
}
\label{fig:degree_impact}
\end{figure*}

\paragraph{Results and Discussion}
\autoref{fig:degree_impact} shows the relationship between the degree of entities and the impact of deleting the knowledge instance related to the entity in the 6-layer model trained on the knowledge graphs.
From \autoref{fig:degree_impact}, it can be seen that, in the case of the ER graph, there is no relationship between the degree of entities and the impact. 
On the other hand, in the case of the BA graph, a relationship between the degree of entities and the impact is observed.
This means that in the model trained on the ER graph, which has structural properties different from the real world, there is no significant difference in the impact of deleting knowledge related to each entity. 
However, in the model trained on the BA graph, which has structural properties closer to the real world, the deletion of knowledge related to entities with higher degrees has a more significant impact. 
In comparison, the deletion of knowledge related to entities with lower degrees has less impact.
Although the absolute value of the impact is small, given the vast amount of knowledge stored in actual language models, the overall impact on the entire knowledge base can be considered significant.
The impact indicates the extent of side effects on other knowledge caused by deleting knowledge related to a specific entity.
Therefore, as our hypothesis, the deletion of important knowledge (related to entities with higher degrees) results in significant side effects. 
In comparison, the deletion of less important knowledge (related to entities with lower degrees) leads to minor side effects.

Additionally, as part of an analysis that includes models with different numbers of layers, \autoref{tab:corr_coef} presents the Pearson correlation coefficient between the degree of entities and their impact in each model trained on the knowledge graphs.
\autoref{tab:impact_stats} shows the statistical values of the impact of entities in each model trained on the knowledge graphs.
From \autoref{tab:corr_coef} and \autoref{tab:impact_stats}, it is observed that for models with 12 and 24 layers, similar to the 6-layer model, the correlation coefficient values between the degree of entities and the impact are small for the ER graph and significant for the BA graph.
This indicates that the behavior of LMs in knowledge deletion varies depending on the structural properties of the knowledge being trained, suggesting the importance of analysis focused on the structural properties of knowledge.

\section{Conclusion}
\label{sec:conclusion}
This paper analyzes the deletion of knowledge in LMs trained on a synthetic knowledge graph, revealing that the side effects of deleting knowledge related to popular entities can be catastrophic.
Furthermore, it is demonstrated that LMs trained on synthetic knowledge graphs recognize paraphrases of entities and relationships, as well as relational knowledge between entities, similar to models trained on real-world knowledge. 
This suggests the effectiveness of studies that employ synthetic knowledge graphs to control the knowledge trained by LMs and to analyze the deletion of knowledge in LMs.

\begin{table}[t]
\centering
\small

\begin{tabular}{ccc}
\toprule
\begin{tabular}[c]{@{}c@{}}Knowledge\\Graph\end{tabular} & Model & \begin{tabular}[c]{@{}c@{}}Correlation\\Coefficient\end{tabular} \\ \midrule \midrule
\multirow{3}{*}{ER} & \phantom{0}6 layers & .2093 \\
 & 12 layers & .2634 \\
 & 24 layers & .3045 \\ \midrule
\multirow{3}{*}{BA} & \phantom{0}6 layers & .9348 \\
 & 12 layers & .9892 \\
 & 24 layers & .7540
 \\
\bottomrule
\end{tabular}
\caption{
Pearson correlation coefficients between the degree of entities and the impact of their deletion in models trained on the knowledge graph.
}
\label{tab:corr_coef}
\end{table}

Future work will extend the analysis of the relationship between the degree of entities involved in the knowledge deletion and the side effects across different structures of synthetic knowledge graphs and knowledge editing methods.
In this study, we utilized ROME, a method of knowledge editing, to analyze the deletion of knowledge. 
Still, since ROME is originally a method of editing knowledge, it is essential to conduct a detailed analysis to see if any traces of the deleted knowledge remain.
Additionally, focusing on the challenges associated with completely deleting knowledge, using synthetic knowledge graphs to control the knowledge in LMs, and analyzing the conditions under which knowledge can be deleted presents a compelling area of research.
Unraveling what constitutes the state of ``knowledge being deleted'' in LMs and how to delete knowledge genuinely is deemed highly significant. 
Furthermore, while we validated using synthetic knowledge graphs in this work, conducting analyses with knowledge graphs represented in natural language is also essential.

\begin{table}[t]
\centering
\small
\begin{tabular}{cccccc}
\toprule
\begin{tabular}[c]{@{}c@{}}Knowledge\\Graph\end{tabular} & Model & Max & Min & Mean & SD \\ \midrule\midrule
\multirow{3}{*}{ER} & \phantom{0}6 layers & 0.20 & 0.00 & 0.05 & 0.04 \\
 & 12 layers & 0.20 & 0.00 & 0.07 & 0.04 \\
 & 24 layers & 0.21 & 0.00 & 0.07 & 0.05 \\ \midrule
\multirow{3}{*}{BA} & \phantom{0}6 layers & 0.60 & 0.00 & 0.05 & 0.08 \\
 & 12 layers & 1.08 & 0.00 & 0.09 & 0.18 \\
 & 24 layers & 1.13 & 0.00 & 0.09 & 0.17
\\
\bottomrule
\end{tabular}
\caption{
Statistical values of the impact $I$ (in units of $10^{-2}$) of deleting a knowledge instance related to an entity in models trained on the knowledge graph.
}
\label{tab:impact_stats}
\end{table}

\newpage

\section*{Acknowledgements}
This work was supported by JSPS KAKENHI Grant Number JP22H00524, JP21K21343, JP21K17814.

\bibliography{custom}

\clearpage
\appendix

\begin{figure}[t]
\centering
\includegraphics[width=\linewidth]{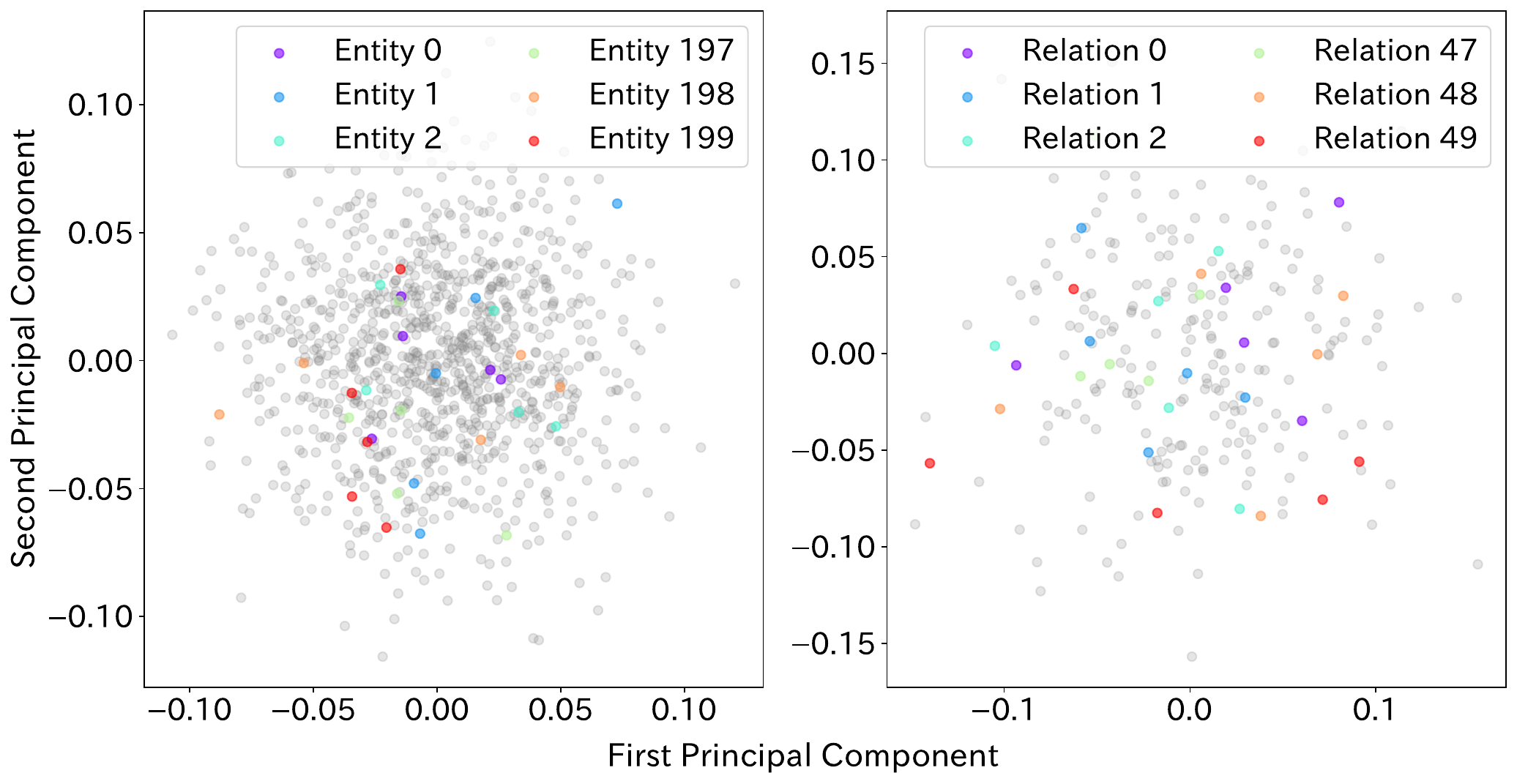}
\caption{
PCA results of the embedding representations in the LM before training the ER graph.
}
\label{fig:emb_before_train}
\end{figure}

\section{Recognition of Paraphrased Representation in LMs}
\label{sec:paraphrase}
As indicated in Section \ref{sec:training}, the synthetic knowledge graphs created in this study possess a structure where each entity and relation has paraphrases, aiming to replicate real-world scenarios more closely.
When training the model on the knowledge graphs with such a structure, we utilized a set of sampled knowledge instances for the training data.
This approach is based on the consideration that if the LM learned all relational knowledge, it might result in the LM merely memorizing each instance of relational knowledge, thereby hindering its generalization ability to recognize paraphrased expressions.

The embedding representations in \autoref{fig:embeddings} from Section \ref{sec:training} show the results of training the model by sampling $20\%$ of all knowledge instances.
It demonstrates that the embedding representations of paraphrases for each entity and relation cluster together.
On the other hand, in the model before training, the embeddings of paraphrases for each entity and relationship are dispersed (\autoref{fig:emb_before_train}). 
This indicates that LMs are not inherently capable of recognizing paraphrases; instead, they acquire the ability to recognize them through learning.

\begin{figure}[t]
\centering
\includegraphics[width=\linewidth]{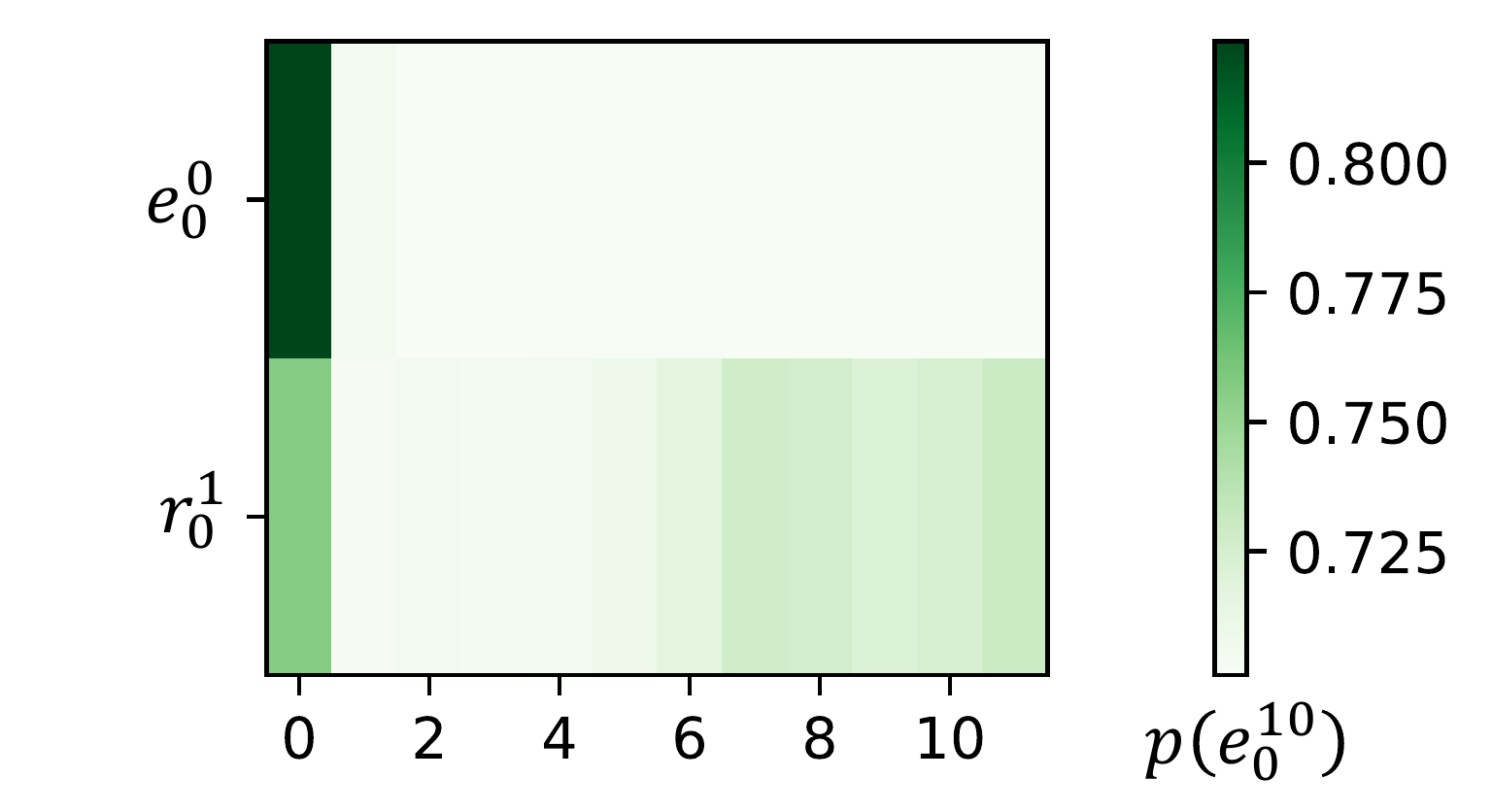}
\caption{
Example of analysis results for the contribution of FF layers in each layer by Causal Tracing (in the case of a 12-layer model).
The horizontal axis represents the layers of the LM, and the vertical axis corresponds to each input token.
The values indicate the difference in the probability of generating the correct token before and after the corrupted-with-restoration run.
The darker areas indicate that restoring the corresponding FF layers allows the model to generate the correct token again, demonstrating that those FF layers contribute to knowledge prediction.
}
\label{fig:causal_tracing}
\end{figure}

\section{Supplemental Information on the Knowledge Editing Method}
\label{sec:causal_tracing}
In Section \ref{sec:ROME}, we briefly introduced ROME, an existing knowledge editing method used in our experiments. This section provides a more detailed explanation of Causal Tracing, a step crucial for locating the parts that play a significant role when the LM associates knowledge.
Causal Tracing analyzes the contribution of each hidden state of the LM during inference through the following procedure\footnote{Manipulations to the hidden states of LMs can also be conceptualized as to the FF layers or attention layers.}.

\begin{enumerate}
    \item \emph{clean run}: First, predictions related to knowledge are made to obtain the normally hidden states of the model. Specifically, all hidden states $\{h_i^{(l)} \,|\, i\in[1,T], l\in[1,L]\}$ are determined when the model predicts $o$ from input $x=(s, r)$.
    Here, $T$ is the length of the input $x$ (in this work, $T=2$), and $L$ is the number of layers in the model.

    \item \emph{corrupted run}: Next, when making predictions related to knowledge, the hidden states of the corrupted model are determined by hiding information about the subject.
    Specifically, when input $x$ is provided, noise is added to the embedding representation $h_1^{(0)}$ corresponding to the subject ($h_1^{(0)} := h_1^{(0)} + \epsilon$).
    Afterward, predictions related to knowledge are made, and the corrupted hidden states $\{h_{i*}^{(l)} \,|\, i\in[1, T], l\in[1, L]\}$ are determined.
    As a result, the correct output that could be output during the clean run can no longer be output during the corrupted run.

    \item \emph{corrupted-with-restoration run}: Finally, for the model with the corrupted hidden states obtained from the corrupted run, specific hidden states $h_{i*}^{(l)}$ are restored to the normally hidden states $h_{i}^{(l)}$ obtained during the clean run.
    This process is performed for each hidden state individually, and predictions related to knowledge are made.
    When the correct output can be output again by restoring a specific hidden state, it indicates that the hidden state contributes to knowledge prediction.
\end{enumerate}

\autoref{fig:causal_tracing} presents an example result of analyzing the contributions of the FF layers at each layer using Causal Tracing.
In this case, the FF layer in the first layer plays a significant role when the LM outputs the correct token $e_0^{10}$ from the input token sequence $(e_0^0, r_0^1)$. 
Thus, Causal Tracing facilitates the identification of parts that play crucial roles in the LM's knowledge prediction.

\end{document}